\documentclass[letterpaper]{article} 
\usepackage[]{aaai23}  
\usepackage{times}  
\usepackage{helvet}  
\usepackage{courier}  
\usepackage[hyphens]{url}  
\usepackage{graphicx} 
\usepackage{natbib}  
\usepackage{caption} 
\usepackage{algorithm}
\usepackage{algorithmic}

\usepackage{newfloat}
\usepackage{listings}
\DeclareCaptionStyle{ruled}{labelfont=normalfont,labelsep=colon,strut=off} 
\floatstyle{ruled}
\newfloat{listing}{tb}{lst}{}
\floatname{listing}{Listing}
\usepackage{subcaption}
\usepackage{booktabs}
\usepackage{multirow}
\usepackage{amsmath}
\usepackage{amssymb}

\nocopyright

\title{Neural Knowledge Bank for Pretrained Transformers}

\author{
    Damai Dai\textsuperscript{1,2}\thanks{~~Joint work of Peking University and Baidu Inc.},
    ~~Wenbin Jiang\textsuperscript{2}, 
    ~~Qingxiu Dong\textsuperscript{1}, \\
    Yajuan Lyu\textsuperscript{2},
    ~~Qiaoqiao She\textsuperscript{2},
    ~~Zhifang Sui\textsuperscript{1} \\
}
\affiliations{
    \textsuperscript{1}MOE Key Lab of Computational Linguistics, Peking University, Beijing, China \\
    \textsuperscript{2}Baidu Inc., Beijing, China \\
    \{daidamai,szf\}@pku.edu.cn,~~dqx@stu.pku.edu.cn \\
    \{jiangwenbin,lvyajuan,sheqiaoqiao\}@baidu.com \\
}

\begin{document}

\maketitle

\begin{abstract}

The ability of pretrained Transformers to remember factual knowledge is essential but still limited for existing models. 
Inspired by existing work that regards Feed-Forward Networks~(FFNs) in Transformers as key-value memories, we design a Neural Knowledge Bank~(NKB) and a knowledge injection strategy to introduce extra factual knowledge for pretrained Transformers. 
The NKB is in the form of additional knowledgeable memory slots to the FFN and the memory-like architecture makes it highly interpretable and flexible. 
When injecting extra knowledge with the Salient Span Masking~(SSM) pretraining objective, we fix the original pretrained model and train only the NKB. 
This training strategy makes sure the general language modeling ability of the original pretrained model is not influenced. 
By mounting the NKB onto the T5 model, we verify its strong ability to store extra factual knowledge based on three closed-book question answering datasets. 
Also, we prove that mounting the NKB will not degrade the general language modeling ability of T5 through two representative tasks, summarization and machine translation. 
Further, we thoroughly analyze the interpretability of the NKB and reveal the meaning of its keys and values in a human-readable way. 
Finally, we show the flexibility of the NKB by directly modifying its value vectors to update the factual knowledge stored in it. 

\end{abstract}

\section{Introduction}

In recent years, large-scale pretrained Transformers~\citep{bert,roberta,unilm,electra,t5} have contributed greatly to natural language processing. 
They are usually trained on large-scale corpora, which contain oceans of factual knowledge. 
When facing some knowledge-intensive downstream tasks such as closed-book question answering, the ability to remember factual knowledge will be especially essential. 
\citet{lama} show that pretrained Transformers can recall some factual knowledge that appears in the training corpus in a zero-shot manner. 
\citet{knowledge_pack} also prove that after fine-tuning, T5~\citep{t5} can answer some open-domain questions without access to external knowledgeable contexts. 
Even so, the ability of pretrained models to store factual knowledge is still limited~\citep{not_kb,guess}. 
Considering this actuality, in this paper, we aim to design an interpretable method to introduce extra factual knowledge for pretrained Transformers. 

\begin{figure}[t]
\centering
\includegraphics[width=0.99\linewidth]{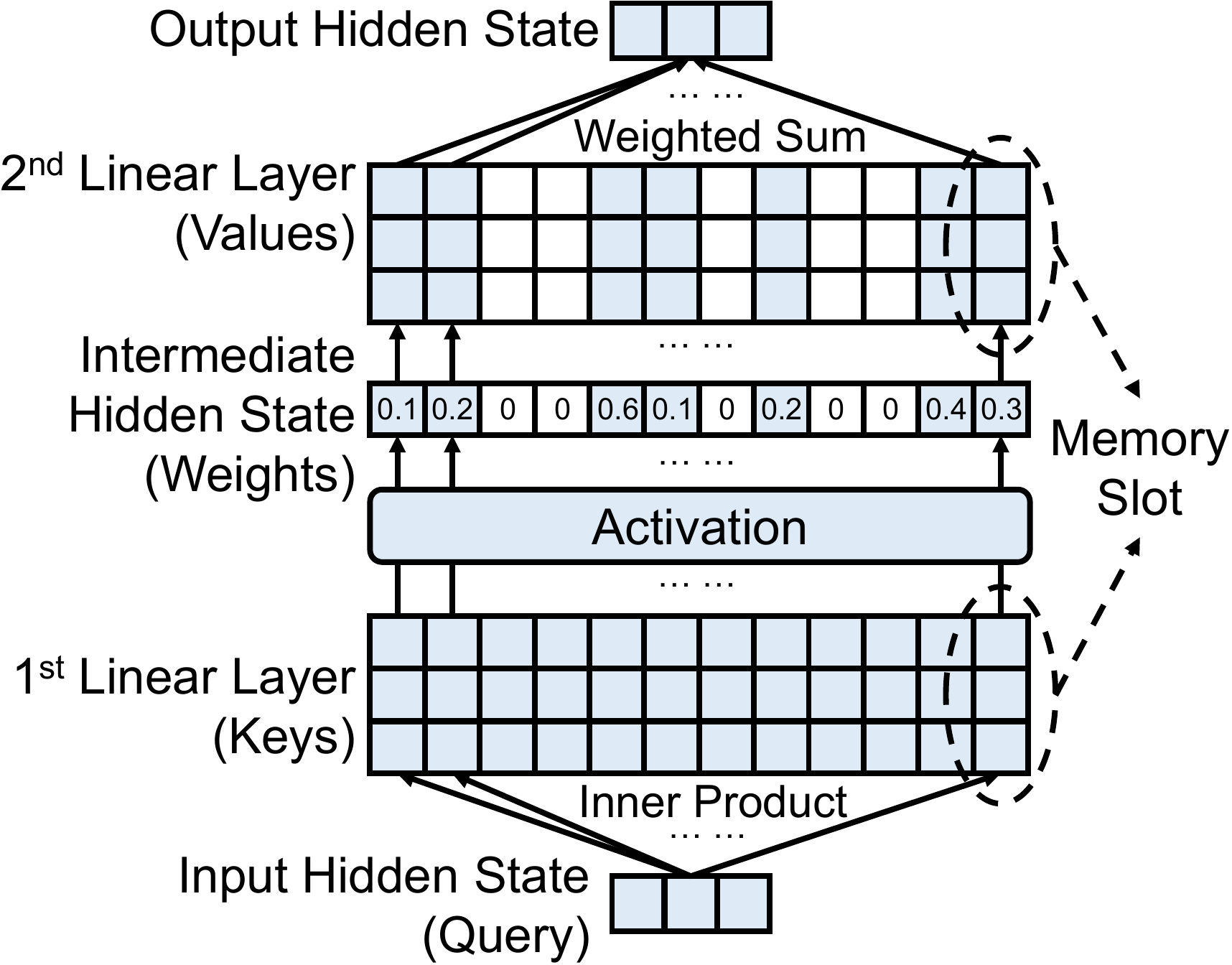}
\caption{
A view of the FFN as key-value memory. 
The first linear layer serves as keys and the second as values. 
Each key-value pair forms a memory slot. 
The intermediate hidden state contains the weights used to integrate the values. 
}
\label{fig:ffn_as_memory}
\end{figure}

\citet{ffn_memory} point out that Feed-Forward Networks~(FFNs) in Transformers work in a similar way to key-value memories. 
As shown in Figure~\ref{fig:ffn_as_memory}, in an FFN, we regard the first linear layer as a series of keys and the second linear layer as the corresponding values. 
The input hidden state of the FFN is fed into the first linear layer and activates a set of intermediate neurons. 
Then, taking these activated neurons as weights, the second linear layer integrates the corresponding value vectors through weighted sum. 
On top of this view, \citet{kn} further find that FFNs in pretrained Transformers store factual knowledge in a memory-like manner. 

Inspired by the above findings, we design a Neural Knowledge Bank~(NKB) and a knowledge injection strategy to introduce extra factual knowledge for pretrained Transformers. 
The NKB is an FFN-like module concatenated after the original FFN as additional knowledgeable memory slots. 
In order to inject factual knowledge, we first acquire a knowledgeable corpus from Wikipedia and then pretrain the NKB with the Salient Span Masking~(SSM)~\citep{realm} pretraining objective. 
Note that during knowledge injection, we fix the original pretrained model to avoid influencing its general language modeling ability. 
For downstream tasks, we can directly fine-tune the whole model. 
The advantages of the NKB are reflected in three aspects: 
(1) The knowledge injection process for the NKB is independent of the original pretrained model, so introducing extra knowledge will not degrade the general language modeling ability of the original model. 
(2) The memory-like architecture of the NKB makes it highly interpretable and we can explain the meaning of its keys and values in a human-readable way. 
(3) The key-value architecture of the NKB has high flexibility and we can easily perform knowledge updating on the NKB by modifying its value vectors. 

On three closed-book question answering datasets in different domains, we find that mounting the NKB can boost the performance of T5, especially in the biomedical domain that the T5 pretraining corpus does not cover much. 
Also, through two representative tasks, summarization and machine translation, we prove that mounting the NKB will not degrade the general language modeling ability of the original T5 model. 
Further, we thoroughly analyze the NKB to reveal its working mechanism and present the meaning of its keys and values in a human-readable way. 
Finally, we show the flexibility of the NKB by directly modifying its value vectors to update the factual knowledge stored in it. 

Our contributions are summarized as follows: 
\begin{itemize}
\item We propose the idea of the NKB, including its architecture and knowledge injection strategy, to neurally store extra factual knowledge for pretrained Transformers. 
\item We verify that the NKB can boost the performance of T5 on closed-book question answering and meanwhile keep the general language modeling ability of the T5 model. 
\item We analyze the interpretability of the NKB and reveal the meaning of its keys and values in a human-readable way.  
\item We show the flexibility of the NKB by directly modifying its value vectors to update the factual knowledge in it. 
\end{itemize}

\begin{figure*}[t]
\centering
\includegraphics[width=0.99\linewidth]{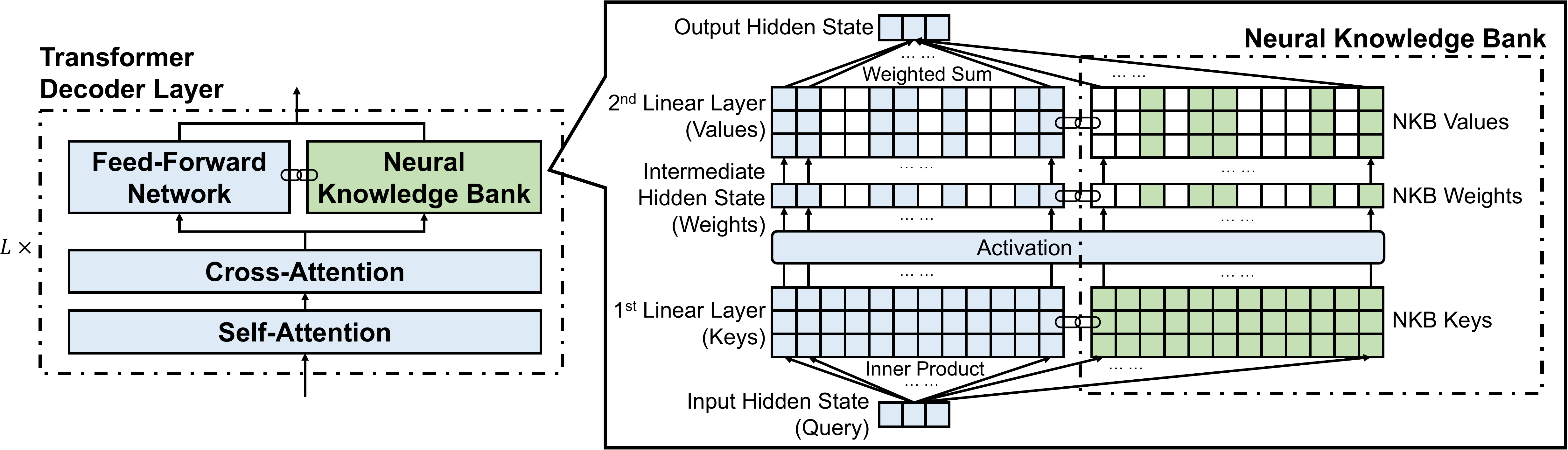}
\caption{
Illustration of the Neural Knowledge Bank~(NKB) for a Transformer decoder layer. 
The NKB shares the same architecture as the FFN and is composed of $d^{\prime}$ additional knowledgeable memory slots. 
During knowledge injection, we freeze the parameters of the original Transformer and inject extra factual knowledge into only the NKB with the Salient Span Masking~(SSM) pretraining objective. 
For downstream tasks, we can directly fine-tune the whole model. 
}
\label{fig:nkb}
\end{figure*}

\section{Background: Transformer}

Recently, Transformer~\citep{transformer} has been the most popular and effective architecture in natural language processing. 
Taking a standard Transformer encoder as an example, it is stacked with $L$ identical Transformer layers. 
In each Transformer layer, there are two main modules: a self-attention~(SelfAtt) module and a feed-forward network~(FFN). 
For an input sequence with $len$ tokens, let $X \in \mathbb{R}^{len \times d}$ denote the input hidden states of a Transformer layer, we formulate these two modules as follows: 
\begin{align}
Q_{h} = X W_{h}^{Q}, \  & K_{h} = X W_{h}^{K}, \  V_{h} = X W_{h}^{V}, \\
\operatorname{SelfAtt}_{h}(X) & = \operatorname{Softmax} \left(Q_{h} K_{h}^{T} \right) V_{h}, \label{equ:self_att} \\
\operatorname{FFN}(H) & = \operatorname{ActFunc} \left(H W_1^{T} \right) W_2, \label{equ:ffn}
\end{align}
where $W_{h}^{Q}, W_{h}^{K}, W_{h}^{V} \in \mathbb{R}^{d \times \frac{d}{n}}, W_1, W_2 \in \mathbb{R}^{4d \times d}$ are parameter matrices, 
$\operatorname{SelfAtt}_{h}(X)$ computes the $h$-th of the $n$ attention heads, 
$H \in \mathbb{R}^{len \times d}$ denotes the output hidden states of the self-attention module, which is computed by projecting the concatenation of all the attention heads, 
and $\operatorname{ActFunc}$ denotes the activation function such as GELU~\citep{gelu} or ReLU~\citep{relu}.
We omit the scaling factor in the self-attention module and the bias terms for simplicity. 

Comparing Equation~(\ref{equ:self_att}) and Equation~(\ref{equ:ffn}), we can find that the calculation formula of $\operatorname{FFN}(\cdot)$ is almost the same as that of $\operatorname{SelfAtt}_h(\cdot)$, except that they have different nonlinear functions. 
Therefore, it is reasonable to view the FFN as a module with the query-key-value mechanism. 
Specifically, the FFN input $H$ serves as queries, the parameters of the first linear layer $W_1$ are keys, and the parameters of the second linear layer $W_2$ are values. 
This view of the FFN is also supported by \citet{ffn_memory,ffn_value,kn}. 

\section{Method}

Following \citep{ffn_memory,ffn_value,kn}, we also view FFNs in Transformer as key-value memories. 
On top of this view, we design a Neural Knowledge Bank~(NKB) and a knowledge injection strategy to introduce extra factual knowledge for pretrained Transformers. 
In this section, we introduce the view of the FFN as key-value memory, the architecture of the NKB, and the knowledge injection method. 

\subsection{Key-value Memory View of FFN}

We first formulate the FFN as key-value memory like \citet{ffn_memory,kn}. 
As illustrated in Figure~\ref{fig:ffn_as_memory}, we regard the input hidden state $\mathbf{h} \in \mathbb{R}^{d}$ of a token as a query, and the parameter matrices $W_1, W_2 \in \mathbb{R}^{4d \times d}$ of two linear layers as $4d$ keys and $4d$ values, respectively, where each key or value is a $d$-dimension vector. 
First, $\mathbf{h}$ is fed into the first linear layer. 
For each key vector in $W_1$, we compute a scalar score $s_i$ through inner product:
\begin{equation}
    s_i = \mathbf{h}^{T} W_1[i, :], 
\end{equation}
where $[\cdot, \cdot]$ denotes the slice operation for a matrix. 
Then, after activation, these scores compose the intermediate hidden state $\mathbf{h}^{(\text{inter})} \in \mathbb{R}^{4d}$ of the FFN:
\begin{align}
    w_i &= \operatorname{ActFunc}(s_i), \\
    \mathbf{h}^{(\text{inter})} &= [w_1; w_2; ...; w_{4d}], 
\end{align}
where $[\cdot; \cdot]$ denotes concatenation. 
Finally, taking these activated neurons as weights, we integrate the value vectors in $W_2$ to get the output hidden state $\mathbf{h}^{(\text{output})} \in \mathbb{R}^{d}$ of the FFN through weighted sum: 
\begin{equation}
    \mathbf{h}^{(\text{output})} = \sum_{i=1}^{4d} w_i W_2[i, :]. 
\end{equation}
In the above formulations, we omit the bias terms. 

\subsection{Neural Knowledge Bank}

Inspired by \citet{kn} who find that FFNs in Transformers can store factual knowledge, we design our NKB with the same architecture as an FFN. 
Specifically, for a Transformer layer, we allocate two new matrices $W_1^{\prime}, W_2^{\prime} \in \mathbb{R}^{d^{\prime} \times d}$ as additional keys and values, where $d^{\prime}$ is a hyper-parameter that control the capacity of the NKB. 
As illustrated in Figure~\ref{fig:nkb}, we mount the NKB onto a Transformer layer by concatenating $W_1^{\prime} \text{ and } W_2^{\prime}$ after $W_1 \text{ and } W_2$ in the original FFN, respectively. 

With the NKB, the intermediate hidden state $\mathbf{h}^{(\text{inter})}$ of the FFN will be extended to $(4d+d^{\prime})$-dimensions: 
\begin{align}
    s_i^{\prime} &= \mathbf{h}^{T} W_1^{\prime}[i, :], \quad w_i^{\prime} = \operatorname{ActFunc}(s_i^{\prime}), \\
    \mathbf{h}^{(\text{inter})} &= [w_1; w_2; ...; w_{4d}; w_1^{\prime}; w_2^{\prime}; ...; w_{d^{\prime}}^{\prime}], 
\end{align}
where $w_i^{\prime}$ is the weight of the $i$-th additional memory slot. 
Finally, taking the new $\mathbf{h}^{(\text{inter})}$ as weights, the value vectors in $W_2$ and $W_2^{\prime}$ are integrated as follows: 
\begin{equation}
    \mathbf{h}^{(\text{output})} = \sum_{i=1}^{4d} w_i W_2[i, :] + \sum_{i=1}^{d^{\prime}} w_i^{\prime} W_2^{\prime}[i, :]. 
\end{equation}
Also, we omit the bias terms in the above formulations. 

As a simple extension of the FFN, the NKB is easy to implement and use. 
More importantly, the memory-like architecture of the NKB makes it highly interpretable and we can explain the meaning of its keys and values in a human-readable way. 
Also, the key-value architecture has high flexibility and we can easily perform knowledge updating on the NKB by directly modifying its value vectors. 

\subsection{Knowledge Injection}

In order to introduce extra factual knowledge, we pretrain the NKB with the Salient Span Masking~(SSM)~\citep{realm,knowledge_pack} pretraining objective. 
To be specific, we first acquire a knowledgeable corpus from Wikipedia using the DrQA~\citep{drqa} document retriever. 
Then, we recognize the salient spans (i.e., named entities and dates) in each text segment using the entity recognizer in spaCy\footnote{https://spacy.io}.
Finally, we randomly mask one of the salient spans and train the parameters of the NKB to reconstruct the masked salient span.

Note that during knowledge injection, we freeze the parameters in the original pretrained model and update only the parameters in the NKB. 
Compared with previous work~\citep{knowledge_pack} that updates the pretrained parameters for knowledge injection, our training strategy can avoid the general language modeling ability of the pretrained model being degraded due to parameter changes. 
In addition, with this training strategy, new factual knowledge is precisely injected into the NKB, so it also becomes easier to locate and analyze the newly introduced factual knowledge. 
After knowledge injection, we can directly fine-tune the whole model for downstream tasks. 

\section{Experiments}

\subsection{Tasks}

Following \citet{knowledge_pack}, we use the closed-book question answering task to evaluate the factual knowledge stored in the parameters of a model. 
Compared with open-domain question answering~\citep{odqa} that requires the model to answer context-independent questions about facts in the real world, the closed-book setting has a more strict constraint that the model cannot access external resources when answering questions. 
Therefore, a model can address closed-book question answering well only if it can store the factual knowledge in its pretrained parameters.

In addition, in order to evaluate the general language modeling ability of a model, we also consider two representative tasks, summarization and machine translation. 
They are not heavily dependent on external factual knowledge and can evaluate whether a model can model and generate languages. 

\subsection{Datasets}
For closed-book question answering, we use three datasets in this paper: two general-domain datasets, Natural Questions~\citep{nq} and WebQuestions~\citep{wq}, and a biomedical-domain dataset, HEAD-QA~\citep{headqa}. 
HEAD-QA is a multiple choice question answering dataset. 
In order to adapt it to our closed-book question answering setting, we extract the correct answer for each question as the output. 
Under the closed-book setting, we ignore all the related documents or contexts for the questions and only retain the questions as input and their corresponding answers as output. 
Following \citet{odqa_lee,odqa_min_a,odqa_min_b,knowledge_pack}, we filter out the examples with answers longer than ﬁve tokens. 
We split the filtered datasets into training and validation sets and the statistics of the datasets are shown in Table~\ref{tab:dataset}. 

For summarization, we use the Xsum~\citep{xsum} dataset. 
For machine translation, we use the English-German translation data in WMT14\footnote{https://www.statmt.org/wmt14}~(WMT-En-De) and English-Romanian translation data in WMT16\footnote{https://www.statmt.org/wmt16}~(WMT-En-Ro). 
For these three datasets, we use the training and validation sets following their official data splits, which are shown in Table~\ref{tab:lm_dataset}.

\begin{table}[t]
\centering
\setlength{\tabcolsep}{20pt}
\begin{tabular}{@{}l c c@{}}
\toprule
\textbf{Dataset}& \textbf{Training} & \textbf{Validation} \\ 
\midrule
Natural Questions & 74,773 & 3,003 \\ 
WebQuestions & ~~3,190 & 1,710 \\
HEAD-QA & ~~~~~436 & ~~~436 \\ 
\bottomrule
\end{tabular}
\caption{
Data splits of three filtered closed-book question answering datasets used in this paper. 
}
\label{tab:dataset}
\end{table}

\begin{table}[t]
\centering
\setlength{\tabcolsep}{24pt}
\begin{tabular}{@{}l c c@{}}
\toprule
\textbf{Dataset}& \textbf{Training} & \textbf{Validation} \\ 
\midrule
Xsum & ~~~204,045 & 11,332 \\ 
WMT-En-De & 4,548,885 & ~~2,169 \\ 
WMT-En-Ro & ~~~610,320 & ~~1,999 \\
\bottomrule
\end{tabular}
\caption{
Official data splits of the summarization and machine translation datasets used in this paper. 
}
\label{tab:lm_dataset}
\end{table}

\begin{table*}[t]
\centering
\setlength{\tabcolsep}{6pt}
\begin{tabular}{@{}l c c c c c c @{}}
\toprule
\textbf{Model} & \textbf{\# Params} & \textbf{\# Knowledgeable Params} & \textbf{Natural Questions} & \textbf{WebQuestions} & \textbf{HEAD-QA} & \textbf{Average} \\ 
\midrule
Transformer & 220M & ~~~N/A & ~~0.4 & ~~2.0 & ~~0.2 & ~~0.9 \\
T5$_{base}$ & 220M & ~~~N/A & 26.3 & 29.9 & ~~3.7 & 20.0 \\ 
T5$_{base}$+NKB-a & 225M & 225M & 26.9 & \textbf{31.9} & \textbf{11.0} & \textbf{23.3} \\ 
T5$_{base}$+NKB & 225M & ~~~~5M & \textbf{27.4} & 31.1 & \textbf{11.0} & \textbf{23.2} \\ 
\midrule
T5$_{large}$ & 770M & ~~~N/A & 28.5 & 31.5 & ~~5.0 & 21.7 \\ 
\bottomrule
\end{tabular}
\caption{
Exact Match~(EM) scores on closed-book question answering. 
\# Knowledgeable Params denote the parameters that are trained during knowledge injection. 
With 5M more parameters, the NKB boosts the performance of T5$_{base}$ by 3.2 EM scores on average, exceeding the performance of T5$_{large}$ with 3.4 times of parameters. 
We also find that during knowledge injection, training only the NKB~(T5$_{base}$+NKB) can achieve a comparable performance with training all the parameters~(T5$_{base}$+NKB-a). 
}
\label{tab:main}
\end{table*}

\subsection{Experimental Settings}

We conduct experiments based on the PyTorch-version HuggingFace \textit{transformers} library\footnote{https://github.com/huggingface/transformers}. 
All the experiments are run on NVIDIA V100 GPUs with 32 GB memory. 

For knowledge injection, we adopt the DrQA~\citep{drqa} document retriever to retrieve the top-3 Wikipedia documents related to each question in a dataset to construct the knowledgeable corpus for the dataset. 
After retrieval, we use the entity recognizer in spaCy to extract named entities and dates for salient span masking. 
We mount the NKB onto the last FFN layer in the decoder of T5$_{base}$~\citep{t5}. 
The number of additional memory slots $d^{\prime}$ is set to 3072, the same as the intermediate hidden dimension in T5$_{base}$. 
When injecting knowledge, we freeze all the parameters in T5$_{base}$ and only update the parameters in the NKB.
We use AdaFactor~\citep{adafactor} as the optimizer and do not apply dropout for the NKB. 
We list the complete hyper-parameters for three datasets in Appendix~\ref{appendix:hyper_know} due to the space limit. 

For downstream tasks, we fine-tune all the parameters in T5$_{base}$+NKB.
We tune the hyper-parameters on the validation set and report the best validation performance for each dataset. 
For closed-book question answering, following~\citep{knowledge_pack}, we use AdaFactor~\citep{adafactor} as the optimizer. 
For summarization and machine translation, we use AdamW~\citep{adamw} as the optimizer. 
The complete hyper-parameters are different for each dataset, so we list them in Appendix~\ref{appendix:hyper_ft} due to the space limit. 

\subsection{Baselines}

In order to show the advantages of the NKB, we compare T5$_{base}$+NKB with four baselines: 
(1) \textbf{Transformer} denotes a vanilla Transformer model that shares the same architecture with T5$_{base}$ but its parameters are randomly initialized without pretraining. 
(2) \textbf{T5$_{base}$} and \textbf{T5$_{large}$} denote the standard T5 pretrained models with 220M and 770M parameters, respectively. 
(3) \textbf{T5$_{base}$+NKB-a} denotes a model that shares the same architecture with T5$_{base}$+NKB but all its parameters are updated during knowledge injection. 

\begin{table*}[t]
\centering
\setlength{\tabcolsep}{19pt}
\begin{tabular}{@{}l c c c c @{}}
\toprule
\textbf{Model}& \textbf{Xsum~(Rouge-L)} & \textbf{WMT-En-De~(BLEU)} & \textbf{WMT-En-Ro~(BLEU)} & \textbf{Average} \\
\midrule
Transformer & 20.7 & 21.5 & 21.9 & 21.4 \\
T5$_{base}$ & \textbf{30.1} & \textbf{30.5} & \textbf{28.2} & \textbf{29.6} \\ 
T5$_{base}$+NKB-a & 24.9 & 25.9 & 25.1 & 25.3 \\ 
T5$_{base}$+NKB & \textbf{30.1} & \textbf{30.5} & \textbf{28.2} & \textbf{29.6} \\ 
\bottomrule
\end{tabular}
\caption{
Results on summarization and machine translation. 
Training all the parameters~(T5$_{base}$+NKB-a) during knowledge injection introduces a negative impact on the performance of T5$_{base}$ for these two tasks. 
By contrast, training only the NKB~(T5$_{base}$+NKB) will not degrade the general language modeling ability of T5$_{base}$. 
}
\label{tab:lm}
\end{table*}

\subsection{Results}

\paragraph{Closed-book Question Answering}
The results on closed-book question answering are shown in Table~\ref{tab:main}. 
We use the Exact Match~(EM) score as the metric, which evaluates whether the generated answer is totally the same as one of the ground-truth answers. 
From the table, we have the following observations: 
(1) The vanilla Transformer without pretraining performs extremely poorly on closed-book question answering since it is not knowledgeable at all. 
(2) With pretraining, T5$_{base}$ achieves a good EM score on the general-domain datasets~(i.e., Natural Questions and WebQuestions), but on the biomedical-domain dataset HEAD-QA, it also performs poorly since biomedical texts account for only a small proportion in the T5 pretraining corpus. 
(3) With more parameters, T5$_{large}$ achieves better performance than T5$_{base}$, but it still cannot address the biomedical-domain closed-book question answering well. 
(4) With only 5M more parameters, the NKB significantly boosts the performance of T5$_{base}$, which approaches the performance of T5$_{large}$ on the general-domain datasets and largely outperforms T5$_{large}$ on the biomedical-domain dataset. 
(5) On closed-book question answering, knowledge injection for only 5M parameters in the NKB~(T5$_{base}$+NKB) can achieve a comparable performance with T5$_{base}$+NKB-a that updates all the parameters during knowledge injection. 

\paragraph{Summarization and Machine Translation}
Ideal knowledge injection should inject new factual knowledge into a pretrained model but not negatively influence the general language modeling ability of the original model. 
We use summarization and machine translation, two representative tasks that do not heavily rely on external factual knowledge to evaluate the general language modeling ability of models. 
We use Rouge-L~\citep{rouge} as the metric for summarization and SacreBLEU~\citep{sacre_bleu} for machine translation. 
We demonstrate the results in Table~\ref{tab:lm}, where the NKB is trained on the WebQuestions-related knowledgeable corpus. 
We have the following findings: 
(1) With pretraining, T5$_{base}$ achieves much better performance than the vanilla Transformer on all the datasets. 
(2) Although performing well on closed-book question answering, T5$_{base}$+NKB-a has a poor performance on summarization and machine translation since it changes the pretrained parameters of T5$_{base}$ during knowledge injection. 
(3) For T5$_{base}$+NKB, we train only the newly introduced parameters, so the general language modeling ability of T5$_{base}$ will not be negatively influenced.  
As a result, T5$_{base}$+NKB can not only address knowledge-intensive tasks like closed-book question answering well, but also keep a good performance on other tasks that do not heavily rely on external factual knowledge. 





\begin{table}[t]
\centering
\setlength{\tabcolsep}{16pt}
\begin{tabular}{@{}l l l@{}}
\toprule
\textbf{Value} & \textbf{Top-scoring Token} & \textbf{Entity Category} \\
\midrule
$\mathbf{v}_{71}$ & Norway & Place \\
$\mathbf{v}_{2878}$ & Constantin & Organization \\
$\mathbf{v}_{2170}$ & Columbus & Person \\
$\mathbf{v}_{221}$ & 1974 & Date \\
$\mathbf{v}_{1581}$ & Portuguese & Other \\
$\mathbf{v}_{1046}$ & desert & Non-entity \\
\bottomrule
\end{tabular}
\caption{
Example NKB value vectors and their top-scoring tokens with the highest probability. 
}
\label{tab:value}
\end{table}

\section{Interpretability of NKB}

\subsection{Value Vectors Store Entities}
\label{sec:value}

\citet{kn} find that FFNs can store factual knowledge, and more specifically, 
\citet{ffn_value} state that the value vectors in FFNs are often corresponding to human-readable concepts. 
Inspired by them, we analyze the NKB value vectors in the output vocabulary space and find that most of the value vectors store specific entities. 

\paragraph{Method}
We randomly sample NKB value vectors from the best T5$_{base}$+NKB checkpoint fine-tuned on WebQuestions for analysis. 
For the $i$-th value vector $\mathbf{v}_i \in \mathbb{R}^{d}$, we first project it into the output vocabulary space and get a probability distribution $\mathbf{p}_i \in \mathbb{R}^{N_{vocab}}$ over the output vocabulary:
\begin{equation}
\mathbf{p}_i = \operatorname{Softmax} (E\mathbf{v}_i),
\end{equation}
where $E \in \mathbb{R}^{N_{vocab} \times d}$ is the output embedding matrix in T5$_{base}$. 
Then, for each sampled value vector $\mathbf{v}_i$, we manually check the token that has the highest probability in $\mathbf{p}_i$ and classify it into one out of six entity categories: \textit{Person}, \textit{Place}, \textit{Organization}, \textit{Date}, \textit{Other}, or \textit{Non-entity}. 
In order to guarantee the confidence of the manual annotation, we ignore tokens that are not complete words or not in English. 
Finally, we annotate 100 valid value vectors. 

\begin{figure}[t]
\centering
\includegraphics[width=0.99\linewidth]{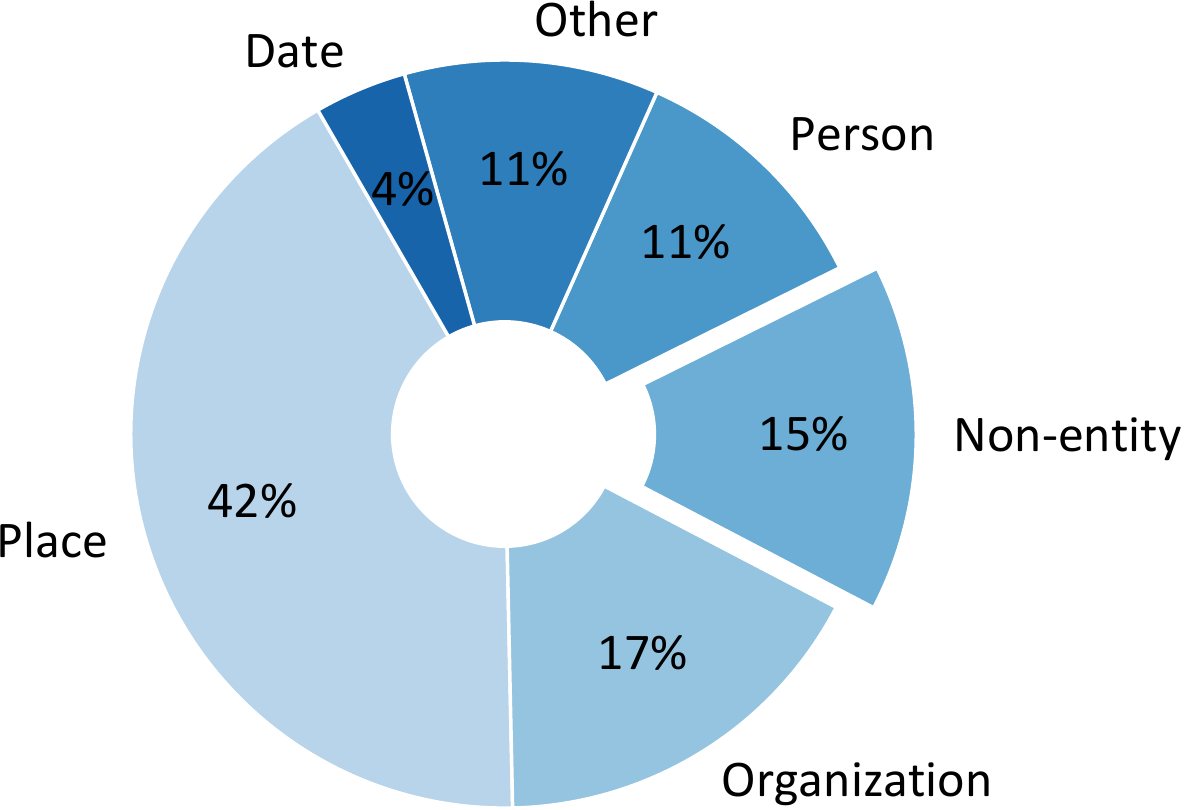}
\caption{
Distribution of the top-scoring tokens for 100 sampled NKB value vectors. 
85\% of the sampled value vectors are corresponding to specific entities. 
}
\label{fig:value}
\end{figure}

\paragraph{Results}
For each entity category, we show an example value vector and its top-scoring token in Table~\ref{tab:value}. 
We also present the distribution of all the top-scoring tokens over the entity categories in Figure~\ref{fig:value}. 
We find that 85\% of the sampled value vectors are corresponding to specific entities. 
The entity information in the value vectors will be integrated into the FFN output hidden state $\mathbf{h}^{(\text{output})}$ and contribute to the generation of the final answer. 

\begin{table*}[t]
\centering
\setlength{\tabcolsep}{33pt}
\begin{tabular}{@{}l l l@{}}
\toprule
\textbf{Key} & \textbf{Top-triggering Questions} & \textbf{Common Semantic Pattern} \\
\midrule
\multirow{3}{*}{$\mathbf{k}_{726}$} & What kind of language do Egyptian speak? & \multirow{3}{*}{Ask about the language of Egypt} \\
 & What was the ancient Egyptians' spoken language? &  \\
 & What language is mainly spoken in Egypt? &  \\
\midrule
\multirow{3}{*}{$\mathbf{k}_{1452}$} & What school did Mark Zuckerberg attend? & \multirow{3}{*}{Ask about the schools of celebrities} \\
 & What college did Albert Einstein go to? &  \\
 & What university did Obama graduate from? &  \\
\midrule
\multirow{3}{*}{$\mathbf{k}_{2750}$} & What style of art did Henri Matisse do? & \multirow{3}{*}{Ask about the art genre} \\
 & What type of artist is Henri Matisse? &  \\
 & What genre of art is the Mona Lisa? &  \\
\bottomrule
\end{tabular}
\caption{
Example NKB key vectors and their top-triggering questions. 
We can identify a common and human-readable semantic pattern among the top-triggering questions for 78\% of the sampled key vectors. 
}
\label{tab:key}
\end{table*}

\subsection{Key Vectors Capture Input Patterns}

Compared with the NKB value vectors that store input-independent entities, the NKB key vectors are input-sensitive. 
They are responsible for determining when to activate the memory slots according to the input patterns. 
In this section, by analyzing when the memory slots are activated, we reveal the patterns that trigger the key vectors. 

\paragraph{Method}
We analyze the same T5$_{base}$+NKB checkpoint used in Section~\ref{sec:value}. 
To be specific, we input each question in the validation set of WebQuestions into the model. 
When the model is generating the first answer token for a question, we record its NKB weights (i.e., the intermediate hidden state in the NKB). 
After the NKB weights for all the questions are recorded, we randomly sample 100 key vectors for analysis. 
For the $i$-th key vector $\mathbf{k}_i \in \mathbb{R}^{d}$, we select out 5 questions that trigger the key vector the most (i.e., have the highest NKB weights). 
Finally, we manually check these 5 questions and attempt to identify human-readable semantic patterns that appear in at least 2 questions. 

\paragraph{Results}
From the manual annotation, we find that 78\% of the sampled keys have common semantic patterns in their top-triggering questions. 
We show some example key vectors and their top-triggering questions in Table~\ref{tab:key} for a better understanding of the patterns.
Putting the findings about the NKB keys and values together, we can summarize the working mechanism of the NKB as follows: the key vectors determine whether to activate the memory slots according to the input patterns, and then the value vectors of the activated memory slots are integrated to contribute to generating the final answer entity.

\begin{table*}[t]
\centering
\setlength{\tabcolsep}{6.5pt}
\begin{tabular}{@{}l l l l@{}}
\toprule
\textbf{Question} & \textbf{Original Prediction} & \textbf{Knowledge Surgery} & \textbf{Updated Prediction} \\
\midrule
What places make up New England? & Maine & $\mathbf{v}_{1112} \text{~+=~} 0.07 (\mathbf{e}_{\text{Europe}} \text{~-~} \mathbf{e}_{\text{Maine}}$) & Europe \\
What are the colors of the NY Giants? & Purple & $\mathbf{v}_{2419} \text{~+=~} 0.07 (\mathbf{e}_{\text{Blue}} \text{~-~} \mathbf{e}_{\text{Purple}}$) & Blue \\
Where do KFC get their chicken from? & Tennessee & $\mathbf{v}_{1716} \text{~+=~} 0.07 (\mathbf{e}_{\text{Kentucky}} \text{~-~} \mathbf{e}_{\text{Tennessee}}$) & Kentucky \\
\bottomrule
\end{tabular}
\caption{
Examples of knowledge updating on the NKB. 
$\mathbf{e}_{\text{x}}$ denotes the output word embedding of the token x. 
By modifying only one value vector, we can update the model prediction to the expected one. 
}
\label{tab:update_example}
\end{table*}

\begin{table}[t]
\centering
\setlength{\tabcolsep}{20pt}
\begin{tabular}{@{}c c c@{}}
\toprule
\textbf{$\lambda$} & \textbf{Success Rate$\uparrow$} & \textbf{Destruction Rate$\downarrow$} \\
\midrule
0.01 & ~~19.4\% & 0.9\% \\
0.03 & ~~49.3\% & 1.5\% \\
0.05 & ~~86.6\% & 1.8\% \\
0.07 & ~~98.5\% & 2.7\% \\
0.09 & 100.0\% & 4.8\% \\
\bottomrule
\end{tabular}
\caption{
Results of knowledge updating. 
$\uparrow$ denotes the higher the better. 
$\downarrow$ denotes the lower the better. 
Keeping a tolerable destruction rate, we can successfully update the factual knowledge for most examples. 
}
\label{tab:update}
\end{table}

\section{Knowledge Updating for NKB}

The knowledge updating ability for a model is meaningful to avoid the model producing erroneous or outdated results. 
Taking advantage of the flexible key-value architecture of the NKB, we propose a simple method to directly update the answer of a question to another one by performing a small knowledge surgery on the NKB. 

\paragraph{Data}
We select examples from the validation set of WebQuestions for knowledge updating. 
In order to guarantee the validity of the target answer, we select the questions for which the model predicts wrong answers, and aim to update these wrong answers to correct ones. 
In this paper, we only consider the questions that have single-token ground-truth answers and predicted answers. 
Finally, we keep 67 examples for the following knowledge updating experiments. 

\paragraph{Method}
Inspired by \citet{kn}, we perform a small knowledge surgery on the NKB to eliminate the information of the original answer and introduce new information of the target answer. 
To be specific, for a question, we first select the memory slot with the highest NKB weight. 
Then, we perform the following operation on the NKB:
\begin{equation}
    \mathbf{v}_{t} \text{~+=~} \lambda (\mathbf{e}_{\text{tgt}} \text{~-~} \mathbf{e}_{\text{ori}}),
\end{equation}
where $t$ is the index of the highest-weight memory slot, $\lambda$ is a hyper-parameter, $\mathbf{e}_{\text{ori}}$ and $\mathbf{e}_{\text{tgt}}$ denote the output word embeddings of the original answer and the target answer, respectively.
Intuitively, by modifying only one value vector~(accounting for only 0.00034\% of the whole model parameters), the knowledge surgery tends to suppress the predicted probability of the original answer and encourage the model to generate the target answer.

\paragraph{Metric}
We use two metrics to evaluate the effects of the knowledge surgery. 
The \textbf{success rate} denotes the proportion of the examples whose answers are successfully updated to the expected ones. 
We also use a \textbf{destruction rate} to measure the influence of the knowledge surgery on other examples. 
For each knowledge surgery, we randomly sample other 5 questions in the validation set of WebQuestions and the destruction rate denotes the proportion of the questions whose answers are changed after the knowledge surgery. 

\paragraph{Results}
We demonstrate the results of knowledge updating in Table~\ref{tab:update}. 
As $\lambda$ becomes larger, the success rate and the destruction rate are both rising. 
Specially, when $\lambda=0.07$, we can achieve a quite high success rate~(98.5\%) while keeping a tolerable destruction rate~(2.7\%). 
In order to better understand the operations and effects of the knowledge surgery, we also present some examples of knowledge updating in Table~\ref{tab:update_example}. 
By modifying only one value vector, we can update the model prediction to the expected one. 

\section{Related Work}

\paragraph{Probing Factual Knowledge in Pretrained Models}
Probing the factual knowledge stored in pretrained models has been a popular topic in recent years. 
Through the fill-in-the-blank cloze task, \citet{lama,lpaqa} find that pretrained models can recall some factual knowledge without fine-tuning. 
\citet{consistency} measure the knowledge consistency of pretrained models using different prompt templates.
\citet{knowledge_pack} fine-tune the T5 model on the closed-book question answering task to evaluate how much knowledge is stored in the model parameters.
\citet{not_kb,guess} also draw some negative conclusions that the factual knowledge in pretrained models is overestimated. 
In summary, pretrained models can store some factual knowledge but to a limited extent, so enhancing pretrained models with extra knowledge is still meaningful. 

\paragraph{Enhancing Pretrained Models with External Knowledge}
Recently, many efforts have been paid to incorporate external knowledge for pretrained models. 
ERNIE~\citep{ernie_tsinghua} and KnowBERT~\citep{know_bert} enhance the word representations with external knowledge graphs during pretraining or continual pretraining.
KEPLER~\citep{kepler} optimizes the objectives of masked language modeling and knowledge embedding jointly on the same pretrained model. 
K-adapter~\citep{kadapter} injects two kinds of knowledge into specific adapters in series while keeping the original pretrained model fixed. 
K-BERT~\citep{kbert} appends relevant triplets to the input sentence to enhance sentence encoding during fine-tuning. 
Kformer~\citep{kformer} retrieves relevant knowledgeable texts during fine-tuning and encodes them into dense vectors to extend the FFN. 
Our NKB combines the advantages of the existing methods: it injects extra knowledge before fine-tuning, can keep the modeling ability of the original pretrained model, and more importantly, is highly interpretable. 

\paragraph{Understanding FFNs in Transformers}

Recently, the Feed-Forward Networks~(FFNs), which account for about two-thirds of the parameters in a Transformer, have been proved especially important~\citep{dynamic_conv,att_is_not_all_you_need} to Transformers. 
\citet{ffn_memory} build a connection between FFNs and key-value memories. 
\citet{ffn_value} further regard the value vectors in FFNs as sub-updates that encode human-readable concepts. 
\citet{kn} point out that FFNs store factual knowledge in a memory-like manner, and the knowledge neurons in FFNs are positively correlated to the expression of their corresponding facts.
Inspired by them, we propose the FFN-like NKB to introduce extra factual knowledge for pretrained Transformers in an interpretable way. 

\section{Conclusion}

In this paper, we propose a Neural Knowledge Bank~(NKB) and a knowledge injection strategy to introduce extra factual knowledge for pretrained Transformers in a neural way.  
On the closed-book question answering task, we verify that the NKB can store extra factual knowledge and thus boost the performance of the T5 model.
Also, we use two representative tasks to prove that mounting the NKB will not degrade the general language modeling ability of the original pretrained model.
Further, we show the interpretability of the NKB by revealing its working mechanism and presenting the meaning of its keys and values in a human-readable way.  
Finally, we show the flexibility of the NKB by directly modifying its value vectors to update the factual knowledge stored in it.
Considering the cost of experiments, we only evaluate and analyze the NKB based on T5 in this paper, but the NKB is also applicable to other pretrained models based on the Transformer architecture. 


\bibliography{custom}

\clearpage
\appendix

\section*{Appendix}

\section{Hyper-parameters for Knowledge Injection}
\label{appendix:hyper_know}

The hyper-parameters for knowledge injection for three closed-book question answering datasets are summarized in Table~\ref{tab:hyper_inj}. 
We train fewer steps for HEAD-QA since the size of its retrieved knowledgeable corpus is relatively small. 

\begin{table*}[ht]
\centering
\setlength{\tabcolsep}{16.3pt}
\begin{tabular}{@{}l c c c@{}}
\toprule
\textbf{Hyper-parameters} & \textbf{Natural Questions} & \textbf{WebQuestions} & \textbf{HEAD-QA} \\
\midrule
\# NKB Slot & 3072 & 3072 & 3072 \\
NKB Position & the last decoder layer & the last decoder layer & the last decoder layer
\\
Dropout for NKB & 0 & 0 & 0 \\
\midrule
Sequence Length & 512 tokens & 512 tokens & 512 tokens \\
Batch Size & 256 & 256 & 256 \\
\midrule
Optimizer & AdaFactor & AdaFactor & AdaFactor \\
Maximum Learning Rate & 1e-3 & 1e-3 & 1e-2 \\
Learning Rate Scheduler & constant with warmup & constant with warmup & constant with warmup \\
Total Steps & 30K & 30K & 3K \\
Warm-up Steps & 5K & 5K & 0.5K \\
\midrule
Gradient Clip Norm & 1.0 & 1.0 & 1.0 \\
Random Seed & 1234 & 1234 & 1234 \\
\bottomrule
\end{tabular}
\caption{
Hyper-parameters for knowledge injection for three closed-book question answering datasets. 
}
\label{tab:hyper_inj}
\end{table*}

\section{Hyper-parameters for Fine-tuning}
\label{appendix:hyper_ft}

The hyper-parameters for closed-book question answering are summarized in Table~\ref{tab:hyper_cbqa}. 
For all the reported models, we use the same hyper-parameters. 

The hyper-parameters for summarization and machine translation are summarized in Table~\ref{tab:hyper_mt_sum}. 
For all the reported models except for the vanilla Transformer, we use the same hyper-parameters. 
For the vanilla Transformer, we train 200K steps for summarization and machine translation since it converges more slowly than pretrained models.

\begin{table*}[ht]
\centering
\setlength{\tabcolsep}{28pt}
\begin{tabular}{@{}l c c c@{}}
\toprule
\textbf{Hyper-parameters} & \textbf{Natural Questions} & \textbf{WebQuestions} & \textbf{HEAD-QA} \\
\midrule
Maximum Sequence Length & 256 tokens & 256 tokens & 256 tokens \\
Batch Size & 768 & 256 & 256 \\
\midrule
Optimizer & AdaFactor & AdaFactor & AdaFactor \\
Maximum Learning Rate & 1e-3 & 1e-3 & 1e-3 \\
Learning Rate Scheduler & constant & constant & constant \\
Total Steps & 10K & 5K & 1K \\
\midrule
Gradient Clip Norm & 1.0 & 1.0 & 1.0 \\
Dropout & 0.1 & 0.2 & 0.2 \\
Dropout for NKB & 0 & 0 & 0 \\
Random Seed & 1234 & 1234 & 1234 \\
\bottomrule
\end{tabular}
\caption{
Hyper-parameters for fine-tuning on closed-book question answering. 
}
\label{tab:hyper_cbqa}
\end{table*}

\begin{table*}[ht]
\centering
\setlength{\tabcolsep}{15pt}
\begin{tabular}{@{}l c c c@{}}
\toprule
\textbf{Hyper-parameters} & \textbf{Xsum} & \textbf{WMT-En-De} & \textbf{WMT-En-Ro} \\
\midrule
Maximum Source Sequence Length & 512 tokens & 128 tokens & 128 tokens \\
Maximum Target Sequence Length & 128 tokens & 128 tokens & 128 tokens \\
Batch Size & 16 & 96 & 96 \\
\midrule
Optimizer & Adam & Adam & Adam \\
Maximum Learning Rate & 3e-5 & 3e-5 & 3e-5 \\
Learning Rate Scheduler & linear with warmup & linear with warmup & linear with warmup \\
Total Steps & 50K & 50K & 50K \\
Warm-up Steps & 10K & 10K & 10K \\
\midrule
Gradient Clip Norm & 1.0 & 1.0 & 1.0 \\
Label Smoothing & 0.1 & 0.1 & 0.1 \\
Beam Size & 4 & 4 & 4 \\
Dropout & 0.1 & 0.1 & 0.1 \\
Dropout for NKB & 0 & 0 & 0 \\
Random Seed & 1234 & 1234 & 1234 \\
\bottomrule
\end{tabular}
\caption{
Hyper-parameters for fine-tuning on summarization and machine translation. 
}
\label{tab:hyper_mt_sum}
\end{table*}

\end{document}